\documentclass[11pt]{article}

\usepackage[preprint]{acl}

\usepackage{times}
\usepackage{latexsym}

\usepackage[T1]{fontenc}

\usepackage[utf8]{inputenc}
\usepackage{pifont}

\usepackage{microtype}

\usepackage{inconsolata}

\usepackage{graphicx}

\usepackage{multirow}
\usepackage{booktabs}
\usepackage{colortbl}

\definecolor{ownrow}{RGB}{210,225,255}

\usepackage{tabularx}              
\usepackage{enumitem}              \usepackage{amsmath}               
\usepackage[breakable]{tcolorbox}  
\usepackage[capitalise,nameinlink,noabbrev]{cleveref}

\crefname{prompt}{Prompt}{Prompts}
\Crefname{prompt}{Prompt}{Prompts}
\definecolor{lightblue}{rgb}{0.93, 0.96, 1.0}
\newcounter{prompt}
\renewcommand{\theprompt}{\arabic{prompt}}
\newcommand{\prompt}[3]{%
  \par\medskip
  \refstepcounter{prompt}%
  \begin{tcolorbox}[colback=lightblue!35, colframe=white!45!black,
                    title={Prompt.~\theprompt:~#1}, breakable]
  #2
  \label{#3}
  \end{tcolorbox}%
  \par\medskip}

\newcommand{\myparagraph}[1]{\vspace{0.3em}\noindent\textbf{#1}}

\title{UniWeaver: Harnessing Unified Multimodal Models for \\ Interleaved Text-Image Generation}

\author{
 \textbf{Hongxu Chen\textsuperscript{1}}\thanks{Equal contribution.},
 \textbf{Zixuan Zhong\textsuperscript{2}\footnotemark[1]},
 \textbf{Donghao Zhou\textsuperscript{3}\footnotemark[1]\thanks{Project lead.}},
 \textbf{Haodong Li\textsuperscript{3}},
\\
 \textbf{Xiaomin Yu\textsuperscript{4}},
 \textbf{Tianhao Li\textsuperscript{5}},
 \textbf{Shiyao Yu\textsuperscript{5}},
 \setcounter{footnote}{3}
 \textbf{Zhongwei Wan\textsuperscript{1}}\thanks{Corresponding authors.},
 \textbf{Pheng-Ann Heng\textsuperscript{3}}\footnotemark[4]
\\
\\
 \textsuperscript{1}OSU,
 \textsuperscript{2}UCL,
 \textsuperscript{3}CUHK,
 \textsuperscript{4}HKUST(GZ),
 \textsuperscript{5}HKU
\\
 \small{
   \textbf{Project Page:} \href{https://poscape.github.io/UniWeaver/}{poscape.github.io/UniWeaver}
 }
}

\begin{document}
\maketitle


\begin{abstract}
Unified multimodal models (UMMs) have opened up new possibilities for generating rich interleaved text-image content. However, they generally lack native support for coordinating the two modalities, and translating coarse user requests into coherent multimodal compositions also requires explicit planning. We propose \textbf{UniWeaver}, a structured, model-agnostic harness framework that integrates multimodal planning, generation, and visual feedback to produce coherent interleaved outputs. To train UMMs for this task, we construct \textbf{UniWeaver-10K}, a dataset of 10,000 tasks across 20 task families, paired with execution trajectories guided by UniWeaver. We also introduce \textbf{UniWeaver-Bench}, a task-oriented benchmark systematically assessing text coverage, writing quality, visual coverage, and image-text alignment. Experiments show that UniWeaver achieves state-of-the-art performance and improves interleaved text-image generation across multiple UMM backbones.

\end{abstract}


\section{Introduction}

Large multimodal models have rapidly improved at understanding and generating visual content. More recently, unified multimodal models (UMMs) have brought visual understanding and generation into a shared architecture \citep{DBLP:conf/cvpr/LuCL0KMHK24,DBLP:journals/corr/abs-2405-09818,DBLP:journals/corr/abs-2409-18869,DBLP:conf/iclr/XieMBZWLGCYS25,DBLP:conf/cvpr/WuCWMLPLXYR025,DBLP:journals/corr/abs-2505-14683}, and related work started to explore the generation of free-form sequences of text and images \citep{DBLP:conf/iclr/DongHPQGYZSZWK024,DBLP:journals/corr/abs-2401-10208,DBLP:conf/cvpr/XingJTMGCZPHXYX26,DBLP:conf/eccv/WangCWWYWZXPWLZ26}. These advances make it possible to move beyond producing solely a single text or image answer toward richer multimodal outputs for travel guides, tutorials, and illustrated explanations \citep{DBLP:conf/emnlp/LiuX0AR0H24,DBLP:conf/cvpr/Zhou0S0XYGZLHZL25}.

\begin{figure}[!t]
 \centering
 \includegraphics[width=\columnwidth]{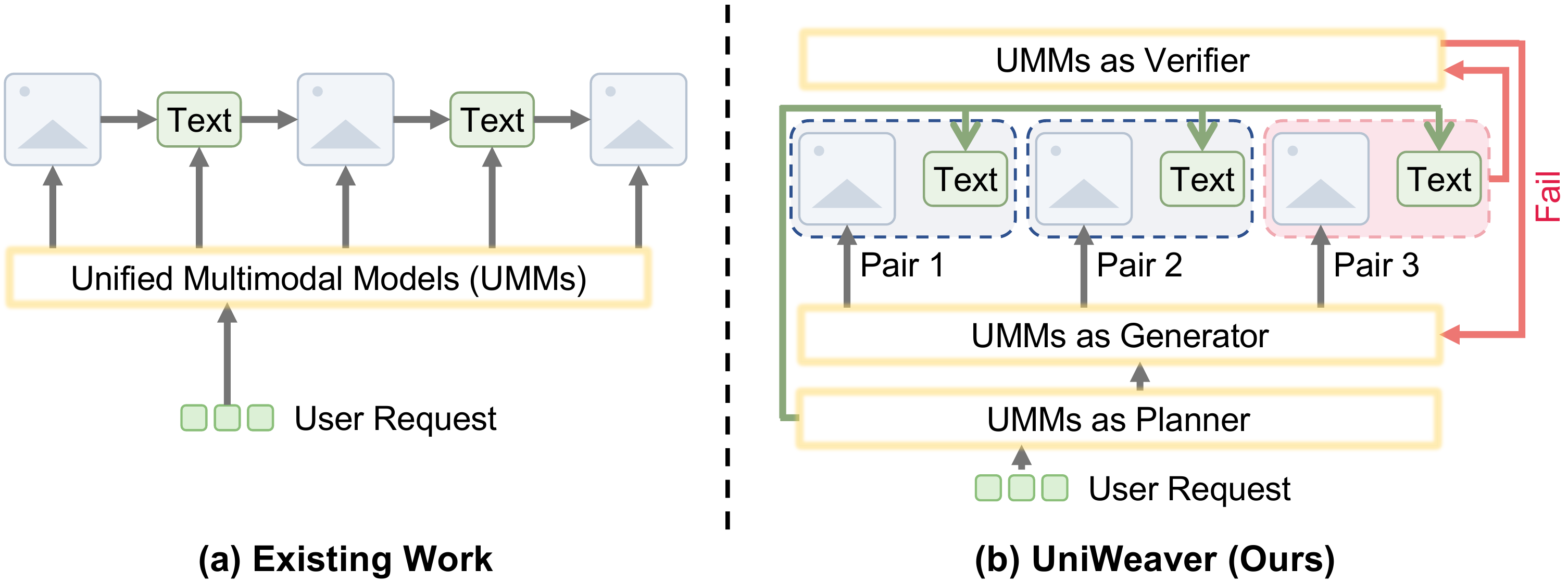}
 \vspace{-6mm}
    \caption{
\textbf{Paradigms of interleaved text-image generation.}
(a) Existing UMMs generate text and images sequentially.
(b) UniWeaver structures multimodal planning, generation, and local visual verification.
}
    \label{fig:paradigm}
\vspace{-4mm}
\end{figure}

However, enabling interleaved text-image generation for UMMs remains challenging. 
First, despite their unified understanding and generation capabilities, UMMs do not generally support the coordinated generation of interleaved text and images as a native capability. Generating such outputs requires not only producing each modality but also coordinating their alternation and ensuring that text and images complement one another throughout the multimodal output. 
Second, translating a user request into meaningful interleaved content requires high-level planning beyond individual text or image generation. User requests typically convey the coarse intended goal without specifying how it should be realized. Bridging this gap requires determining the granularity and number of content units, allocating information across text and images, and organizing them into a coherent sequence that jointly fulfills the user's intent.

\begin{figure*}[h]
    \centering
    \includegraphics[width=\textwidth]{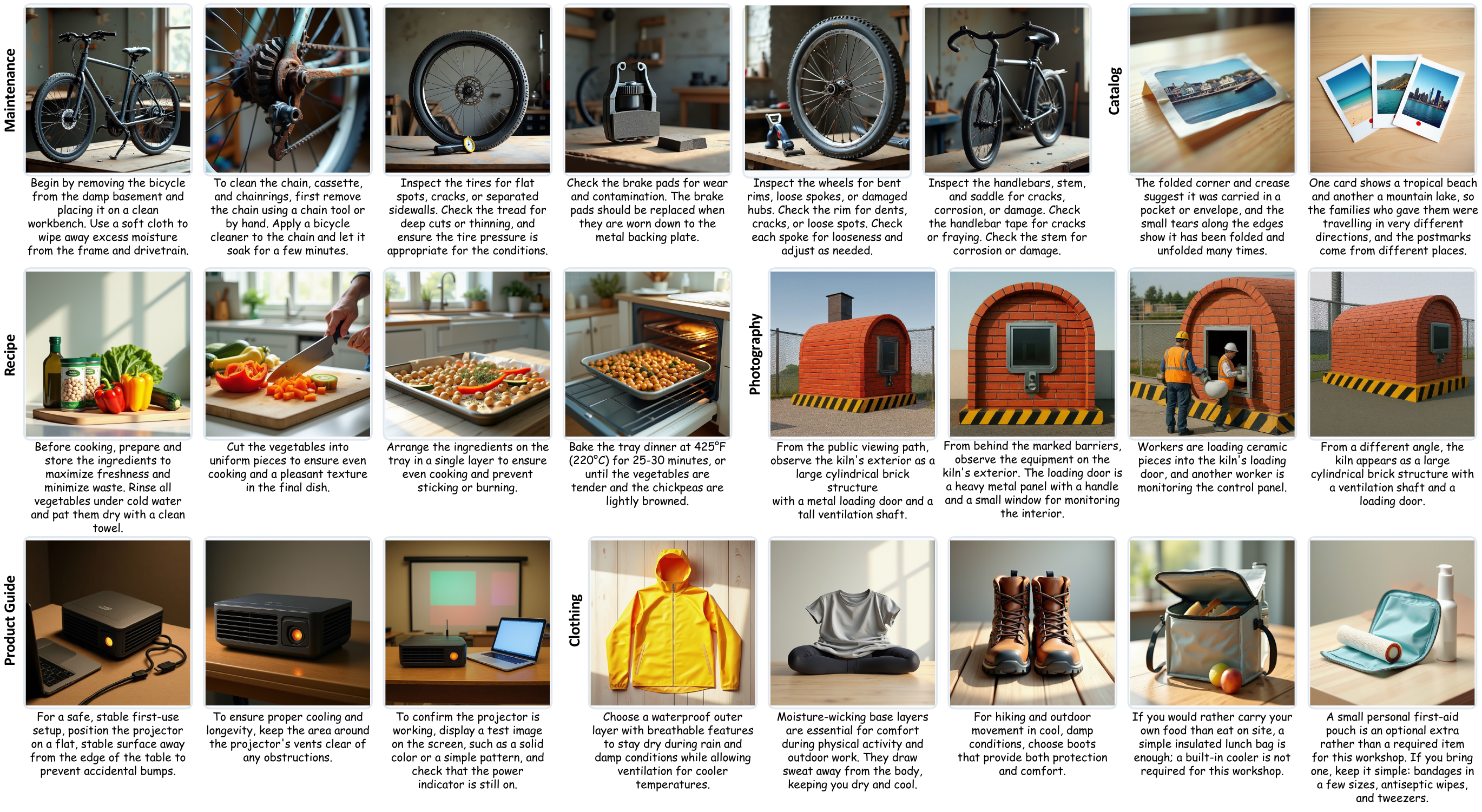}
    \vspace{-8mm}
    \caption{
\textbf{Generated results of the proposed UniWeaver.}
Our harness framework enables coherent interleaved text-image generation
for UMMs across diverse real-world tasks, producing task-relevant
visuals paired with informative and actionable text.
}
\vspace{-4mm}
    \label{fig:teaser}
\end{figure*}

To address these challenges, we propose \textbf{UniWeaver}, a structured harness framework that integrates multimodal planning, generation, and visual feedback into a unified workflow (Figure~\ref{fig:paradigm}). UniWeaver decomposes a user request into content units, determines what should be conveyed through text or images, and organizes them into a coherent interleaved output. This process coordinates the roles of the two modalities, ensuring that text and images complement each other rather than being generated as isolated outputs. By explicitly planning the content and structure before generation, UniWeaver bridges the gap between coarse user requests and complete, coherent multimodal outputs. It then produces reader-facing prose and image-generation instructions, optionally maintains reusable visual references across images, and verifies generated images through observable visual checks. Images that fail verification can be refined independently, reducing unnecessary sequential dependencies between unrelated content units. It is worth noting that the framework's design is model-agnostic and can be instantiated with different UMM backbones.
Figure~\ref{fig:teaser} provides several examples of UniWeaver's
generated results, illustrating its ability to produce coherent
and informative multimodal content.

To train UMMs to plan and generate interleaved text-image content from natural user requests, we construct \textbf{UniWeaver-10K}, a specific dataset from 10,000 tasks spanning 20 task families. For each task, we use a strong LLM guided by the UniWeaver framework to construct a corresponding execution trajectory, yielding 10,000 execution trajectories for training. 
We then use these trajectories to train UMMs to improve their ability to autonomously plan and coordinate interleaved text-image generation.
By deriving these trajectories from user requests that specify goals rather than predefined multimodal structures, our data construction process provides dense supervision for learning how to organize content and coordinate text-image generation within the framework. 

Moreover, evaluating whether a generated interleaved output truly fulfills a user's request requires more than assessing its individual modalities or their shallow coherence. 
Existing benchmarks such as InterleavedBench \citep{DBLP:conf/emnlp/LiuX0AR0H24}, ISG-Bench \citep{DBLP:conf/iclr/ChenCPLWCLH00K25}, and OpenING \citep{DBLP:conf/cvpr/Zhou0S0XYGZLHZL25} measure only text quality, visual quality, and other limited properties, but these properties alone do not ensure that the output covers the information and visual evidence required by the user. 
We therefore introduce \textbf{UniWeaver-Bench}, a task-oriented benchmark that directly evaluates the completed outputs of reader-facing text and generated images. It measures task fulfillment through text coverage, writing quality, visual coverage, and image-text alignment, enabling comprehensive comparisons across methods with different generation mechanisms.
Experiments demonstrate that our approach achieves state-of-the-art performance on UniWeaver-Bench and consistently improves interleaved text-image generation when integrated with multiple UMM backbones.

Our main contributions are as follows:
\begin{itemize}
    \setlength{\itemsep}{0pt}
    \setlength{\topsep}{0pt}
    \vspace{-4pt}
    \item We develop \textbf{UniWeaver}, a structured, model-agnostic harness framework for UMMs, which integrates multimodal planning, generation, and visual feedback for interleaved text-image generation.

    \item  We construct \textbf{UniWeaver-10K}, a dataset of 10,000 tasks spanning 20 task families, paired with execution trajectories for training UMMs to plan and generate coherent interleaved text-image content.

    \item We introduce \textbf{UniWeaver-Bench}, a task-oriented benchmark for the systematic evaluation of interleaved text-image generation.
\end{itemize}


\section{Related Work}

\myparagraph{Unified Multimodal Models.}
Unified multimodal models (UMMs) aim to support visual
understanding and generation within a single model \citep{xie2026reconstruction}.
Unified-IO 2 \citep{DBLP:conf/cvpr/LuCL0KMHK24} unifies multiple input and output modalities
through a shared semantic space,
while Chameleon \citep{DBLP:journals/corr/abs-2405-09818} and Emu3\citep{
DBLP:journals/corr/abs-2409-18869} mixed image-text sequences
autoregressively.
Show-o \citep{DBLP:conf/iclr/XieMBZWLGCYS25} and Janus
\citep{DBLP:conf/cvpr/WuCWMLPLXYR025} explore different architectures for
unifying visual understanding and generation.
More recently, BAGEL \citep{DBLP:journals/corr/abs-2505-14683} advances unified multimodal
pretraining for understanding, generation, editing,
and multimodal reasoning.
Wan-Weaver \citep{DBLP:conf/cvpr/XingJTMGCZPHXYX26} further explores interleaved generation
through a planner-visualizer architecture and
decoupled training for textual planning and visual
consistency modeling.
These features provide capable foundations
for multimodal understanding and generation.

\myparagraph{Multimodal Generation Systems.}
Related work has explored controllable visual generation
\citep{zhou2024magictailor,liu2026hifi,zhou2026omnishow,xie2026paint}
and consistent visual storytelling
\citep{zhou2026identitystory,song2025scenedecorator}.
Planning and feedback have been widely studied for
solving multi-step tasks
\citep{DBLP:conf/iclr/YaoZYDSN023,
DBLP:conf/nips/ShinnCGNY23,zhou2026metapoint}.
These ideas have also been extended to multimodal
settings, where models can reason over visual
observations \citep{zhou2026thinkv2v} and interact with external tools
\citep{DBLP:journals/corr/abs-2303-11381}.
For interleaved generation, ISG-Agent introduces
a plan--execute--refine pipeline that coordinates
multimodal tools
\citep{DBLP:conf/iclr/ChenCPLWCLH00K25},
while InterleaveThinker employs planner and critic
agents to guide image generators through global
planning and step-wise visual refinement
\citep{DBLP:journals/corr/abs-2606-13679}.
These studies demonstrate the role of planning and
feedback in coordinating multimodal generation processes.
UniWeaver integrates these capabilities into a structured
harness that decomposes user requests into content units,
coordinates the complementary roles of text and images,
and supports visual verification and targeted refinement.
The resulting execution trajectories also provide
supervision for training UMMs to plan and generate
interleaved content.

\myparagraph{Interleaved Generation and Evaluation.}
Interleaved generation models have explored producing
free-form sequences of text and images.
GILL \citep{DBLP:conf/nips/KohFS23} connects language models with image generation
modules,
while DreamLLM and MM-Interleaved \citep{DBLP:conf/iclr/DongHPQGYZSZWK024,
DBLP:journals/corr/abs-2401-10208} investigate
the generation of interleaved multimodal sequences.
Alongside advances in generation, benchmarks have also
been developed to assess the quality of complete
multimodal outputs \citep{huang2026mmdeepresearch}.
InterleavedBench \citep{DBLP:conf/emnlp/LiuX0AR0H24} evaluates text quality, perceptual
quality, image coherence, text-image coherence,
and helpfulness.
ISG-Bench \citep{DBLP:conf/iclr/ChenCPLWCLH00K25} introduces finer-grained evaluation of
interleaved outputs,
while OpenING \citep{DBLP:conf/cvpr/Zhou0S0XYGZLHZL25} expands evaluation to diverse
open-ended real-world tasks and introduces IntJudge
for multimodal assessment.
More recently, WeaverBench \citep{DBLP:conf/cvpr/XingJTMGCZPHXYX26} evaluates prompt adherence,
narrative coordination, content consistency, image
consistency, and completeness.
Building on these evaluation efforts, UniWeaver-Bench
provides a task-oriented assessment of completed
interleaved outputs through text coverage, writing
quality, visual coverage, and image-text alignment.
It evaluates final reader-facing content,
thereby enabling comparison across multimodal generation workflows.


\section{Methodology}

\begin{figure*}[t]
    \centering
    \includegraphics[width=\textwidth]{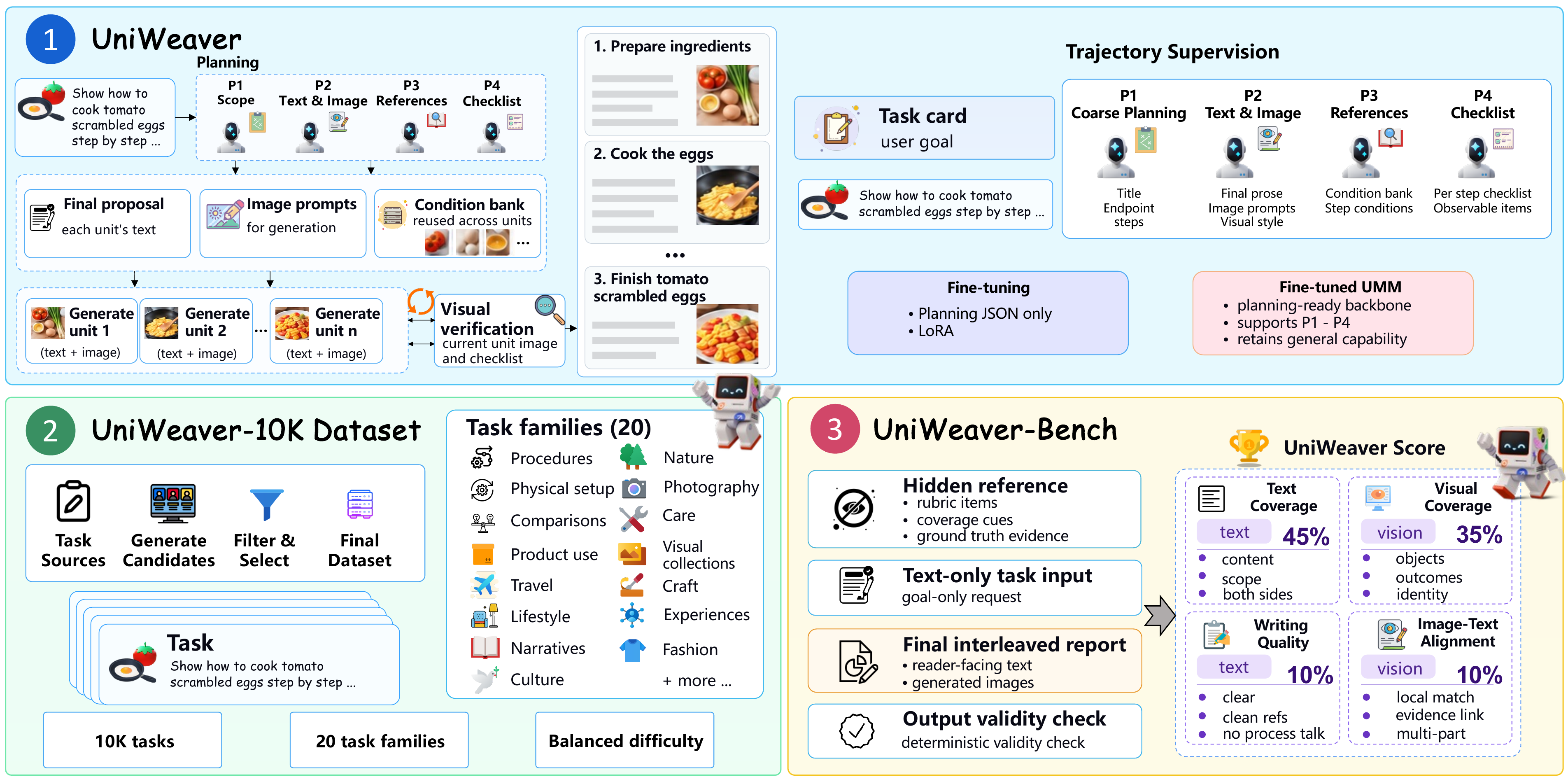}
    \vspace{-6mm}
    \caption{
    \textbf{Overview of UniWeaver.}
    \ding{172} \textit{UniWeaver} introduces a structured workflow for interleaved text-image generation, with execution trajectories providing supervision for fine-tuning (Section~\ref{sec:3.1}).
    \ding{173} \textit{UniWeaver-10K Dataset} provides 10,000 goal-only tasks paired with structured planning trajectories across 20 task families (Section~\ref{sec:3.2}).
    \ding{174} \textit{UniWeaver-Bench} systematically evaluates final interleaved outputs on 120 held-out tasks (Section~\ref{sec:3.3}).
    }
    \vspace{-3mm}
    \label{fig:framework}
\end{figure*}

We introduce \textbf{UniWeaver} (Section~\ref{sec:3.1}), a structured, model-agnostic harness framework for interleaved text-image generation from natural user requests. As shown in Figure~\ref{fig:framework}, UniWeaver integrates multimodal planning, generation, selective visual conditioning, and visual verification into a unified workflow, enabling coherent coordination between text and images. To support training and systematic evaluation, we further construct \textbf{UniWeaver-10K} (Section~\ref{sec:3.2}), a dataset of 10,000 tasks spanning 20 task families, paired with execution trajectories for structured planning supervision, and \textbf{UniWeaver-Bench} (Section~\ref{sec:3.3}), a 120-task benchmark for evaluating the quality and task fulfillment of final interleaved outputs.

\subsection{Overall Framework of UniWeaver}
\label{sec:3.1}

As illustrated by Figure~\ref{fig:framework}, UniWeaver converts a natural request into a structured multimodal plan, realizes the planned text-image units, and verifies generated visual content before assembling the final response.
Let $x$ denote a natural user request. UniWeaver produces an interleaved output:
\begin{equation}
\mathcal{Y}
=
\left[
(c_1,I_1),\ldots,(c_N,I_N)
\right],
\end{equation}
where $N$ denotes the number of content units, and $c_i$ and $I_i$ represent the reader-facing text and generated image of the $i$-th unit, respectively.

\myparagraph{Multimodal Planning.}
Users typically describe a high-level task goal rather than a predefined structure, leaving the number of content units, the output organization, and the allocation of information between text and images unspecified. 
UniWeaver therefore performs multimodal planning before generation to determine these structures and establish a coherent organization for the final output, which consists of four stages:
\vspace{-6pt}
\begin{enumerate}[label=(\arabic*), leftmargin=*]

    \setlength{\itemsep}{-2pt}
    \setlength{\topsep}{-8pt}
    
    \item \textit{{Scope Decomposition.}}
    UniWeaver first constructs a coarse content structure:
    \begin{equation}
    P_1
    =
    \left(
    \tau,e,\{(s_i,g_i)\}_{i=1}^{N}
    \right),
    \end{equation}
    where $\tau$ is the output title, $e$ is the task endpoint, and each unit contains a visual subject $s_i$ and textual content goal $g_i$. The endpoint defines what constitutes a complete response, while the unit-level goals specify what each part should cover.

    \item \textit{{Text-Visual Realization.}}
    Given the content structure, UniWeaver produces reader-facing text and image-generation instructions:
    \begin{equation}
    P_2
    =
    \left(
    v,\{(p_i,c_i)\}_{i=1}^{N}
    \right),
    \end{equation}
    where $p_i$ is the image prompt, $c_i$ is the reader-facing text, and $v$ denotes a shared visual style. This separation allows explanatory or relational information to remain in text while the image prompt focuses on directly visualizable content. The shared style provides a lightweight global constraint across generated images.
    
    \item \textit{{Selective Visual Conditioning.}}
    When a visual concept, such as a specific object, should remain consistent across units, UniWeaver constructs an optional reference bank and routes only the relevant references to each unit. We denote the routed reference set for unit $i$ as $R_i$. Unlike fully sequential generation, where each image may depend on the preceding image, UniWeaver introduces such dependencies only when they are useful, allowing unrelated units to remain independent.

    \item \textit{{Visual Verification.}}
    For each unit, UniWeaver derives a compact checklist $C_i$ of directly observable visual requirements, such as the presence of an object, action, state change, or visible outcome. These checks are restricted to properties that can be judged from the image itself and do not depend on unconstrained and hidden factual reasoning.

\end{enumerate}
\vspace{-4pt}
\myparagraph{Generation and Refinement.}
As shown in Figure~\ref{fig:framework}, UniWeaver first generates any references required by Stage 3 and then produces each image from its planned prompt, shared style, and routed references:
\begin{equation}
I_i^{(0)}
=
G(\tilde{p}_i,R_i),
\qquad
\tilde{p}_i=p_i\oplus v,
\end{equation}
where $G$ denotes the visual generation capability of the underlying backbone.
Each image is then checked against its unit-specific checklist:
\begin{equation}
V(I_i^{(r)},C_i)
\rightarrow
(j_i^{(r)},f_i^{(r)}),
\end{equation}
where $j_i^{(r)}$ is the verification result and $f_i^{(r)}$ provides feedback for failed criteria. If verification fails, the feedback is appended to the original generation instruction and only that unit is regenerated. The framework allows at most two additional attempts after the initial generation, while successful units remain unchanged. Verification therefore acts as a local repair mechanism rather than requiring regeneration of the entire response.

\myparagraph{Supervised Fine-Tuning.}
As shown in the first part of Figure~\ref{fig:framework}, we use the structured outputs from UniWeaver-10K to adapt the planning and understanding pathway of the unified backbone. Unlike approaches that train separate planning and critic models \citep{DBLP:journals/corr/abs-2606-13679}, UniWeaver directly fine-tunes the unified backbone while keeping its visual generation pathway frozen.
For each task $x$, the supervision target is
\begin{equation}
Y=(P_1,P_2,P_3,P_4),
\end{equation}
and the model is trained autoregressively using
\begin{equation}
\scalebox{0.9}{$\displaystyle
\mathcal{L}_{\mathrm{SFT}}
=
-\sum_{k=1}^{4}
\sum_{t=1}^{|P_k|}
\log
p_{\theta}
\left(
P_{k,t}
\mid
x,P_{<k},P_{k,<t}
\right).
$}
\end{equation}
The loss is applied only to the generated planning text tokens. For instance, with BAGEL \citep{DBLP:journals/corr/abs-2505-14683} as the backbone, LoRA \citep{DBLP:conf/iclr/HuSWALWWC22} is applied to the understanding decoder pathway while the image-generation-specific experts remain frozen. This allows the model to internalize the UniWeaver planning procedure while preserving its pretrained visual synthesis capability. Detailed optimization settings are provided in Section~\ref{sec:experimental_setup}.

\subsection{UniWeaver-10K Dataset}
\label{sec:3.2}

\begin{figure*}[t]
    \centering
    \includegraphics[width=\textwidth]{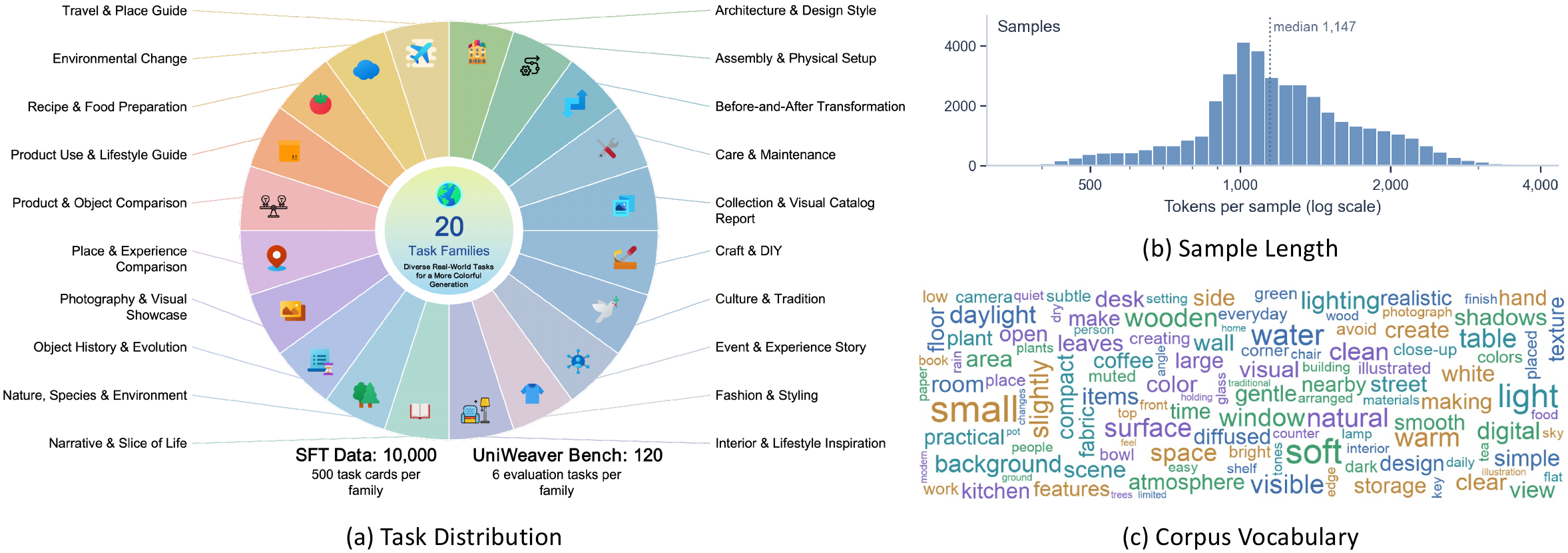}
    \caption{
\textbf{Statistics of UniWeaver-10K and UniWeaver-Bench.}
(a) Task distribution across 20 diverse families, with balanced
coverage in both datasets.
(b) Distribution of training sample lengths in tokens.
(c) Vocabulary characteristics of the training corpus,
visualized as a word cloud.
}
    \label{fig:dataset_statistics}
\end{figure*}

The second part of Figure~\ref{fig:framework} illustrates
the construction of UniWeaver-10K through candidate generation,
filtering, and selection. UniWeaver-10K contains 10,000 tasks evenly distributed across 20 task families,
with 500 tasks per family. The selected tasks follow an
approximately $1{:}2{:}1$ easy-medium-hard distribution.
Figure~\ref{fig:dataset_statistics} further presents dataset statistics, highlighting the diversity of tasks and their characteristics.

\myparagraph{Trajectory Generation.}
We define task families covering procedures, physical setup, transformations, comparisons, product use, travel, lifestyle, narratives, culture, architecture, nature, photography, and related multimodal scenarios. Using DeepSeek-v4-Pro \citep{DBLP:journals/corr/abs-2606-19348} as the teacher model, we generate more than 30,000 candidate requests from family-specific combinations of topics, audiences, situations, constraints, and intended outcomes.
A key design choice is that each request specifies only the user goal. We remove prompts that prescribe a fixed image count, storyboard, output outline, or explicit text-image correspondence. The model must therefore infer how the final response should be organized rather than recover that structure from the input.

\myparagraph{Filtering and Supervision.}
Candidates are filtered for task-family consistency, naturalness, multimodal usefulness, format validity, and duplication, followed by stratified selection over family and difficulty. For each task, DeepSeek-v4-Pro generates the same four-stage structure used at inference time. Later stages are conditioned on earlier ones to maintain cross-stage consistency.
We retain examples that satisfy the UniWeaver planning schema and consistency checks. Supervision consists only of the structured planning outputs; generated image pixels and explicit chain-of-thought are not used as training targets. 
We obtain 10,000 structured execution trajectories for training UMMs to plan and coordinate interleaved text-image generation.

\subsection{UniWeaver-Bench}
\label{sec:3.3}

Figure~\ref{fig:framework}
illustrates UniWeaver-Bench, which evaluates final
interleaved outputs through a validity check and four criteria: text coverage,
writing quality, visual coverage, and image-text alignment.

\myparagraph{Test Samples.}
As shown in Figure~\ref{fig:dataset_statistics}(a), it contains 120 held-out tasks across the same 20 task families and 4 major categories, with 6 tasks per family and 40 easy, 40 medium, and 40 hard examples.
To ensure a reliable evaluation of generalization, benchmark tasks are held out from training, with overlap prevented through task identity and normalized prompt hash matching.
Each task provides only a natural-language goal and is paired with hidden evaluation references containing task-specific coverage requirements and observable visual evidence. A deterministic validity check first verifies that the output contains usable text-image units with one-to-one image pairing.

\myparagraph{Evaluation Metrics.}
Valid outputs are evaluated along four dimensions: \textit{text coverage (TC)}, \textit{writing quality (WQ)}, \textit{visual coverage (VC)}, and \textit{image-text alignment (ITA)}. Each criterion is rated as \emph{not met}, \emph{partially met}, or \emph{met}, mapped to $0$, $0.5$, and $1$. The quality of a gate-passing task $t$ is 
\begin{equation}
\scalebox{0.8}{$\displaystyle
Q_t
=
0.45\,\mathrm{TC}_t
+
0.10\,\mathrm{WQ}_t
+
0.35\,\mathrm{VC}_t
+
0.10\,\mathrm{ITA}_t
$}.
\end{equation}
To account for failures to produce a valid report, we compute for each family $f$
\begin{equation}
S_f=\bar{Q}_f\sqrt{r_f},
\qquad
\mathrm{Total}
=
\frac{1}{20}\sum_{f=1}^{20}S_f,
\end{equation}
where $\bar{Q}_f$ is the mean of $Q_t$ over the gate-passing tasks in family $f$ and $r_f$ is its gate pass rate. Thus, invalid outputs are penalized without allowing report availability to completely dominate the quality of successful generations. We independently evaluate each output using three judges: Kimi K3 \citep{DBLP:journals/corr/abs-2607-24653}, DeepSeek-V4.1-Flash \citep{DBLP:journals/corr/abs-2609-19969}, and GLM-5.3-Flash \citep{DBLP:journals/corr/abs-2602-15763}. Each judge performs two evaluations at temperature $0$,
with the median retained.
Cross-judge ratings are conservatively aggregated,
with disagreements resolved toward the lower tier (Appendix~\ref{sec:apx-eval}).


\section{Experiments}

\subsection{Experimental Setup}
\label{sec:experimental_setup}

We evaluate seven UMM backbones: BAGEL \citep{DBLP:journals/corr/abs-2505-14683}, OmniGen2 \citep{DBLP:conf/cvpr/WuWZYXLWLJLZXLL26}, Show-o2 \citep{DBLP:conf/nips/XieYS25}, TokenFlow \citep{DBLP:conf/cvpr/QuZLWJGYDYW25}, ILLUME+ \citep{DBLP:journals/corr/abs-2504-01934}, Janus-Pro \citep{DBLP:journals/corr/abs-2501-17811}, and SEED-X \citep{DBLP:journals/corr/abs-2404-14396}. The baseline employs a minimal step-wise harness,
generating successive text-image pairs conditioned on
the user request and preceding steps, without explicit
planning or visual verification.
UniWeaver applies its four-stage framework to the same
backbones without updating model parameters. For supervised adaptation, we fine-tune BAGEL on
UniWeaver-10K using LoRA while keeping its image-generation
experts frozen, denoted as UniWeaver-SFT.
We use the AdamW optimizer with a learning rate of
$5\times10^{-5}$, LoRA rank $32$, alpha $64$, and
dropout $0.05$, training for $40{,}000$ steps (two epochs).
Additional optimization details are provided in
Appendix~\ref{sec:apx-training}. All methods are evaluated on UniWeaver-Bench using
the metrics defined in Section~\ref{sec:3.3}.

\begin{table*}[t]
\centering
\caption{
\textbf{Quantitative Comparison on the 120-task UniWeaver-Bench.}
Each major category contains 5 task families covering 30 tasks.
The best result is \textbf{bold} and the second best \underline{underlined}.
\textcolor{red!55!black}{Red} values indicate performance gains over the corresponding UMM baseline. Only gate-passing tasks are averaged, and a criterion whose judge medians span more than one tier scores $0$. Each family's score is multiplied by $\sqrt{r_f}$, where $r_f$ is its gate pass rate.
}
\vspace{-2mm}
\label{tab:main_results}

\newcommand{\gain}[1]{%
    \textcolor{red!55!black}{\textbf{\textit{$\Delta\,#1$}}}%
}
\newcommand{\uwname}{\text{UniWeaver}}
\newcommand{\uwbackbone}[1]{\hspace{1em}\small + #1}

\newcommand{\gcell}[2]{%
    \begin{tabular}[t]{@{}c@{}}
        #1\\[-2pt]
        {\tiny #2}
    \end{tabular}%
}
\newcommand{\uwmodel}[1]{%
    \shortstack[l]{%
        UniWeaver\\[-2pt]
        \hspace{1em}\small + #1%
    }%
}

\small
\setlength{\tabcolsep}{3pt}
\renewcommand{\arraystretch}{1.5}
\setlength{\extrarowheight}{1pt}

\resizebox{\textwidth}{!}{%
\begin{tabular}{l|ccccc|ccccc|ccccc|ccccc|ccccc}
\specialrule{1.2pt}{0pt}{0pt}
\multirow{2}{*}{\textbf{Model}}
& \multicolumn{5}{c|}{\textbf{Procedures \& Instructions}}
& \multicolumn{5}{c|}{\textbf{Comparison \& Recommendation}}
& \multicolumn{5}{c|}{\textbf{Observation \& Description}}
& \multicolumn{5}{c|}{\textbf{Explanation \& Storytelling}}
& \multicolumn{5}{c}{\textbf{Overall}} \\
\cline{2-6}\cline{7-11}\cline{12-16}\cline{17-21}\cline{22-26}
& TC & WQ & VC & ITA & Total
& TC & WQ & VC & ITA & Total
& TC & WQ & VC & ITA & Total
& TC & WQ & VC & ITA & Total
& TC & WQ & VC & ITA & Total \\
\hline

BAGEL
& 14.8 &  7.3 & 48.0 & 14.1 & 25.6
& 28.3 & 14.2 & 35.1 & 14.1 & 27.8
& 26.1 &  4.9 & 45.9 & 15.0 & 29.8
& 16.3 &  6.8 & 52.2 & 13.0 & 27.6
& 21.4 &  8.3 & 45.3 & 14.0 & 27.7 \\

OmniGen2
& 12.3 & 13.8 & 47.9 & 21.9 & 25.9
& 19.6 & 14.2 & 40.4 & 24.7 & 26.8
& 16.0 &  2.5 & 44.6 & 23.3 & 25.4
& 23.2 &  5.4 & 53.3 & 40.4 & 33.7
& 17.8 &  9.0 & 46.5 & 27.6 & 27.9 \\

TokenFlow
& 30.2 & 24.6 & \underline{65.4} & 40.3 & 43.0
& 32.1 &  8.8 & 55.8 & 25.8 & 37.4
& 31.3 &  6.7 & 60.4 & 27.2 & 38.6
& 28.1 & 10.4 & 60.3 & 29.4 & 37.7
& 30.4 & 12.6 & 60.5 & 30.7 & 39.2 \\

ILLUME+
& 23.3 & 27.1 & 51.7 & \underline{47.2} & 36.0
& 35.2 & 10.8 & 55.0 & 35.0 & 39.7
& 34.2 & 12.9 & 57.1 & 26.5 & 39.3
& 27.1 &  5.8 & 66.7 & 25.8 & 38.7
& 30.0 & 14.2 & 57.6 & 33.6 & 38.4 \\

Show-o2
& 10.0 & 43.4 & 21.5 &  9.2 & 17.3
& 15.6 & 36.6 & 36.7 & 12.5 & 24.8
& 20.1 & 30.2 & 38.4 & 13.2 & 26.8
& 19.3 & 32.5 & 41.7 & 21.8 & 28.7
& 16.2 & 35.7 & 34.6 & 14.2 & 24.4 \\

Janus-Pro
& 18.8 &  9.5 & 27.5 &  5.1 & 19.5
& 22.7 &  6.6 & 24.0 &  6.4 & 19.9
& 23.0 &  6.2 & 45.2 &  9.8 & 27.8
& 10.8 &  4.2 & 24.9 &  6.2 & 14.6
& 18.8 &  6.6 & 30.4 &  6.9 & 20.5 \\

SEED-X
&  2.7 &  1.7 & 25.8 &  3.9 & 10.8
&  4.6 &  1.2 & 29.6 &  0.0 & 12.5
&  2.7 &  0.0 & 32.5 &  0.8 & 12.7
&  0.0 &  0.0 & 32.2 &  0.0 & 11.3
&  2.5 &  0.7 & 30.0 &  1.2 & 11.8 \\
\hline

UniWeaver
& 47.3 & 90.1 & 60.3 & 44.1 & 55.8
& 47.1 & \underline{91.9} & \underline{60.0} & \underline{50.9} & 56.5
& 45.1 & \underline{86.4} & \underline{66.2} & \underline{56.7} & \underline{57.8}
& 56.0 & \underline{89.4} & 72.8 & 58.4 & \underline{65.5}
& 48.9 & \underline{89.4} & \underline{64.8} & \underline{52.5} & 58.9 \\[-4pt]
\hspace{1em}\small + BAGEL
& & & & & {\tiny \gain{+30.2}}
& & & & & {\tiny \gain{+28.6}}
& & & & & {\tiny \gain{+28.0}}
& & & & & {\tiny \gain{+37.9}}
& & & & & {\tiny \gain{+31.2}} \\

UniWeaver
& 42.3 & 73.1 & 58.8 & 45.8 & 51.5
& 41.1 & 65.5 & 49.9 & 33.9 & 45.9
& 39.2 & 74.2 & 61.1 & 48.6 & 51.3
& 52.9 & 64.6 & \underline{73.5} & \underline{61.6} & 62.2
& 43.9 & 69.4 & 60.8 & 47.5 & 52.7 \\[-4pt]
\hspace{1em}\small + OmniGen2
& & & & & {\tiny \gain{+25.6}}
& & & & & {\tiny \gain{+19.1}}
& & & & & {\tiny \gain{+25.9}}
& & & & & {\tiny \gain{+28.5}}
& & & & & {\tiny \gain{+24.8}} \\

UniWeaver
& \underline{68.1} & \underline{92.8} & 56.0 & 40.9 & \underline{63.6}
& \underline{58.9} & 86.5 & 47.4 & 50.4 & \underline{56.8}
& \underline{53.3} & 74.6 & 59.1 & 54.9 & 57.6
& \underline{62.6} & 83.2 & 63.0 & 56.0 & 64.2
& \underline{60.7} & 84.3 & 56.4 & 50.5 & \underline{60.5} \\[-4pt]
\hspace{1em}\small + TokenFlow
& & & & & {\tiny \gain{+20.7}}
& & & & & {\tiny \gain{+19.3}}
& & & & & {\tiny \gain{+19.0}}
& & & & & {\tiny \gain{+26.4}}
& & & & & {\tiny \gain{+21.3}} \\

\rowcolor{ownrow}
UniWeaver
& \textbf{86.9} & \textbf{98.3} & \textbf{95.8} & \textbf{79.2} & \textbf{90.4}
& \textbf{81.7} & \textbf{97.5} & \textbf{89.2} & \textbf{80.6} & \textbf{85.8}
& \textbf{78.1} & \textbf{100.0} & \textbf{90.0} & \textbf{86.7} & \textbf{85.3}
& \textbf{81.5} & \textbf{95.8} & \textbf{90.7} & \textbf{88.3} & \textbf{86.8}
& \textbf{82.0} & \textbf{97.9} & \textbf{91.4} & \textbf{83.7} & \textbf{87.1} \\[-4pt]
\rowcolor{ownrow}
\hspace{1em}\small + BAGEL-SFT
& & & & & {\tiny \gain{+64.8}}
& & & & & {\tiny \gain{+57.9}}
& & & & & {\tiny \gain{+55.5}}
& & & & & {\tiny \gain{+59.2}}
& & & & & {\tiny \gain{+59.4}} \\

\specialrule{1.2pt}{0pt}{0pt}
\end{tabular}%
}
\vspace{-4mm}
\end{table*}

\subsection{Main Results}

\textbf{Quantitative Comparison.}
As shown in Table~\ref{tab:main_results}, minimal step-wise harness with UMM backbones yields limited performance on open-ended interleaved tasks, with BAGEL, OmniGen2, and TokenFlow achieving overall scores of 27.7, 27.9, and 39.2, respectively. In contrast, UniWeaver substantially improves their performance to 58.9, 52.7, and 60.5 without parameter updates. These consistent gains demonstrate the effectiveness of structured multimodal planning and generation across different backbones.
Training with UniWeaver-10K further improves performance.
UniWeaver-SFT achieves the highest overall score of 87.1, substantially outperforming both framework-only
UniWeaver (58.9) and minimal step-wise harness (27.7) on BAGEL.
This demonstrates the additional benefit of learning structured planning and generation from execution trajectories.
The improvements extend across multiple evaluation dimensions.
On BAGEL, UniWeaver increases TC from 21.4 to 48.9, WQ from 8.3 to 89.4, VC from 45.3 to 64.8, and ITA from 14.0 to 52.5.
UniWeaver-SFT further achieves 82.0 TC, 97.9 WQ, 91.4 VC, and 83.7 ITA. These results highlight improvements in task-relevant coverage, reader-facing writing quality, visual support, and text-image coordination.

\begin{figure*}[t]
    \centering
    \includegraphics[width=\textwidth]{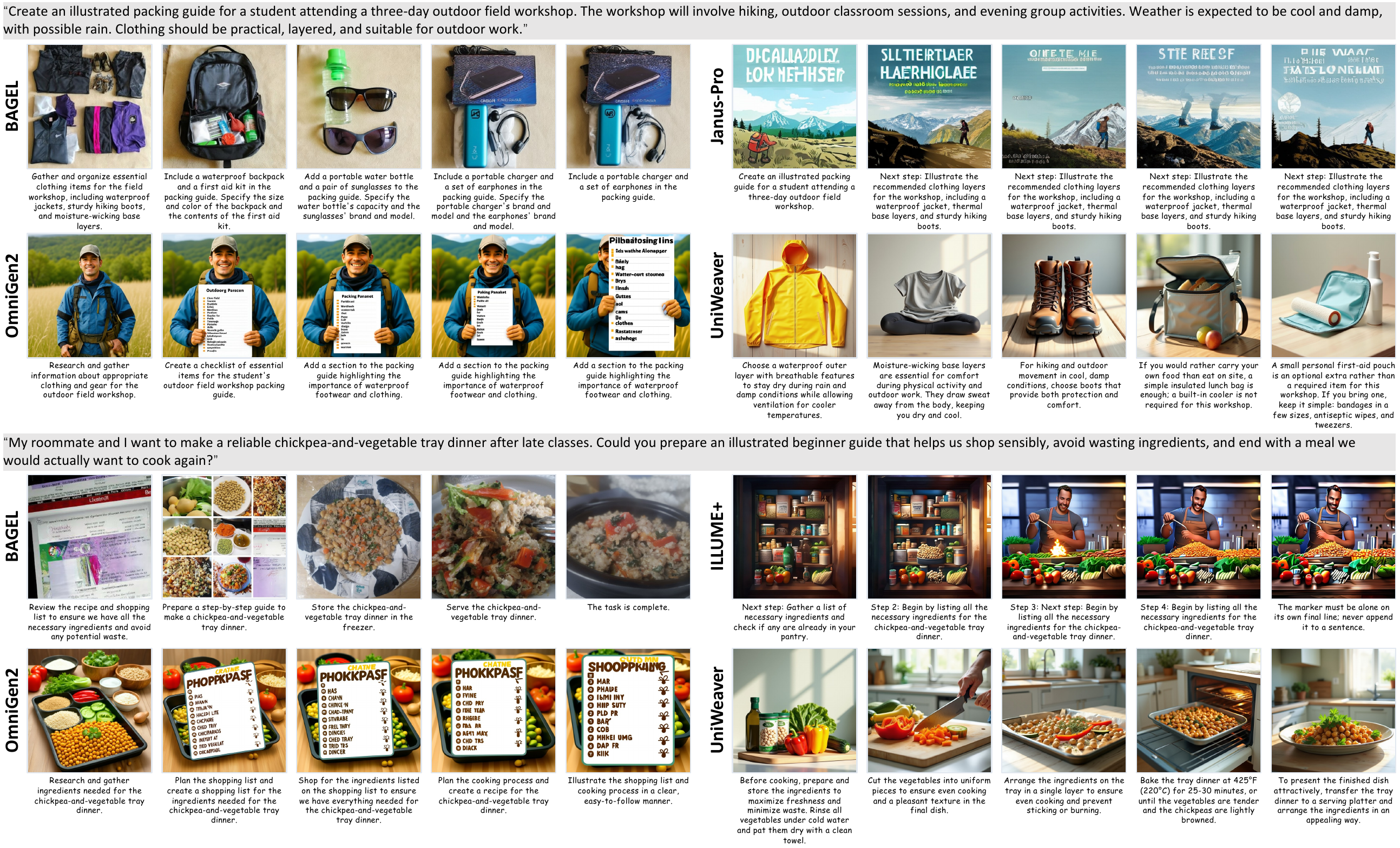}
    \vspace{-6mm}
    \caption{
\textbf{Qualitative comparison on two representative tasks.}
UniWeaver produces more coherent, actionable instructions with task-relevant visuals, while baselines exhibit repetitive content, missing details, or ineffective visual support.
\textit{Zoom in for better view.}
}
\vspace{-3mm}
    \label{fig:qualitative_comparison}
\end{figure*}

\myparagraph{Qualitative Comparison.}
Figure~\ref{fig:qualitative_comparison} compares UniWeaver with competitive baselines on two representative tasks: an outdoor workshop packing guide and a beginner cooking guide, examining their ability to produce coherent, informative interleaved outputs. The baselines frequently exhibit repetitive content, incomplete guidance, and ineffective visual support. For the packing guide,
BAGEL introduces less relevant accessories, OmniGen2 repeats checklist illustrations, and Janus-Pro generates illegible text within images. For the cooking guide, BAGEL omits key steps,
OmniGen2 produces unreadable shopping lists, and ILLUME+ repeats similar scenes and instructions. 
In contrast, UniWeaver delivers practical, well-organized guidance with clearer, task-relevant visuals. The packing guide provides specific clothing and equipment recommendations with explanations of their practical benefits and corresponding illustrations.
The cooking guide progresses coherently from preparation to baking and serving, providing actionable details such as cooking temperature and duration, with images depicting distinct stages of the process.
These results highlight UniWeaver's advantages in task-relevant coverage, writing quality, visual effectiveness, and text-image alignment. Rather than merely producing related text and images,
UniWeaver organizes complementary multimodal content into coherent, actionable guides that better fulfill user requests.

\subsection{Ablation Studies}

We conduct seven ablation studies on UniWeaver + BAGEL to assess the contributions of key components of the proposed framework, as shown in
Table~\ref{tab:ablation}. Specifically, $P_1$ examines task-level and unit-level goals, $P_2$ evaluates shared visual style and text-image role separation, and $P_3$ investigates reference selection, routing, and conditioning.

\begin{table}[t]
\centering
\small
\setlength{\tabcolsep}{6pt}
\renewcommand{\arraystretch}{1.12}
\caption{
\textbf{Ablation studies of UniWeaver.}
All variants are re-run on the same 120 tasks with the same evaluation setup.
Note that $\Delta$ denotes the performance gap measured against the full method.
}
\vspace{-2mm}
\label{tab:ablation}
\begin{tabularx}{\columnwidth}{@{}Xrr@{}}
\toprule
\textbf{Ablation} & \textbf{Total} & \textbf{$\Delta$} \\
\midrule
UniWeaver (w/ BAGEL backbone)             & \textbf{58.89} & --- \\
\addlinespace[2pt]
\multicolumn{3}{@{}l@{}}{%
    \begingroup
    \setlength{\fboxsep}{1pt}%
    \colorbox{gray!20}{%
        \makebox[\dimexpr\columnwidth-2\fboxsep\relax][l]{%
            \textbf{$\boldsymbol{P_1}$: Task Scope}%
        }%
    }%
    \endgroup
} \\
\hspace{1em}w/o Task Goal             & 53.01 & $-$5.88 \\
\hspace{1em}w/o Per-step Content Goal           & 55.56 & $-$3.33 \\
\multicolumn{3}{@{}l@{}}{%
    \begingroup
    \setlength{\fboxsep}{1pt}%
    \colorbox{gray!20}{%
        \makebox[\dimexpr\columnwidth-2\fboxsep\relax][l]{%
            \textbf{$\boldsymbol{P_2}$: Text and Image Planning}%
        }%
    }%
    \endgroup
} \\
\hspace{1em}w/o Shared Visual Style            & 48.21 & $-$10.68 \\
\hspace{1em}w/ Merged Image Prompt and Content         & 53.48 & $-$5.41 \\
\multicolumn{3}{@{}l@{}}{%
    \begingroup
    \setlength{\fboxsep}{1pt}%
    \colorbox{gray!20}{%
        \makebox[\dimexpr\columnwidth-2\fboxsep\relax][l]{%
            \textbf{$\boldsymbol{P_3}$: Visual References}%
        }%
    }%
    \endgroup
} \\
\hspace{1em}References Fixed before the Text        & 57.68 & $-$1.21 \\
\hspace{1em}w/o Per-step Reference Routing      & 57.89 & $-$1.00 \\
\hspace{1em}w/o All Reference Images             & 58.48 & $-$0.41 \\
\bottomrule
\end{tabularx}
\vspace{-2mm}
\end{table}

\myparagraph{Observation \#1: Explicit task goals anchor goal-driven content planning.}
Removing the task goal reduces the overall score from 58.89 to 53.01 (-5.88), while removing per-step content goals yields 55.56 (-3.33). The larger degradation without the task goal highlights the importance of defining what constitutes a complete response. Meanwhile, unit-level goals help specify what each part should contribute. Together, these results support planning content around both the overall user objective and the responsibilities of individual units.

\myparagraph{Observation \#2: Coherent interleaved generation requires visual consistency and distinct modality roles.}
Removing the shared visual style causes the largest performance drop among all ablations, reducing the score to 48.21 (-10.68). Combining reader-facing content and image instructions into a single field also reduces performance, yielding a score of 53.48 (-5.41). These results highlight the importance of maintaining a consistent visual presentation across content units and explicitly distinguishing information to be communicated through text from content to be depicted in images.

\myparagraph{Observation \#3: Selective visual conditioning supports cross-unit coherence.}
Fixing references before text planning reduces the score from 58.89 to 57.68 (-1.21), while removing per-step reference routing or all reference images yields 57.89 (-1.00) and 58.48 (-0.41), respectively.
These results suggest that selecting references with access to the planned image instructions, together with unit-specific reference routing, helps coordinate visual content across units.


\section{Conclusion}

In this work, we study interleaved text-image generation from natural user requests. We introduce UniWeaver, a model-agnostic harness that structures open-ended requests into multimodal content units, coordinates text and image generation, and enables selective visual conditioning and local verification. We further develop UniWeaver-10K for trajectory supervision and UniWeaver-Bench for evaluating task fulfillment from final outputs. Experiments across multiple multimodal backbones show that UniWeaver improves generation without parameter updates, while supervised fine-tuning yields further gains. 
These findings suggest that effective interleaved generation requires not only multimodal generation capabilities, but also structured planning and coordination to fulfill user intent through coherent multimodal outputs.


\section*{Limitations}

Despite its effectiveness, UniWeaver remains constrained by the capabilities of underlying UMMs, whose generated text and images may not yet meet professional-grade quality standards for demanding applications. In addition, our current study focuses on a defined set of interleaved generation tasks, leaving broader task scenarios and application settings to be explored. Future work will investigate stronger multimodal backbones and extend UniWeaver to a wider range of tasks and applications, further improving the quality and applicability of interleaved multimodal generation.

\section*{Ethical Considerations}

UniWeaver builds on existing UMMs and therefore inherits their potential biases and limitations in generated text and images. Although the framework improves the organization and consistency of multimodal outputs, it does not independently guarantee the factual accuracy, fairness, or appropriateness of all generated content. In practical applications, users should exercise appropriate judgment when using generated outputs, particularly in contexts where accuracy and reliability are important. Future work may incorporate additional safeguards and more comprehensive evaluations to support the responsible deployment of interleaved multimodal generation systems.

\bibliography{custom}


\clearpage
\appendix

\section{Implementation and Method Details}
\label{sec:apx-system}

UniWeaver executes four planning stages before rendering the final interleaved output. The executor first generates reference images, then renders planned unit images and pairs them with reader-facing text.
Visual verification supports targeted regeneration of failed images, with at most two additional attempts per unit.

\subsection{Stages and Processes}

The four planning stages execute sequentially, with earlier decisions remaining fixed. Each stage uses a predefined JSON schema, and subsequent stages operate on the preceding planning outputs rather than generated image pixels.
All stage calls use a fixed system instruction requiring schema-compliant JSON output without additional commentary.
The prompts below specify the behavior of the individual stages, with placeholders filled at inference time; inference appends a JSON-only closing instruction, while API-based construction sets JSON response mode.

\myparagraph{Plan \#1: Scope and Sequence.} Decomposes the user request into content units, specifying the task endpoint and each unit's visual subject and reader-facing content goal.

\prompt{Scope and Sequence}{%

Task request: \texttt{\{task\}}

Use task details, ordinary knowledge, and plausible scene detail. Keep an unspecified subject at the category level instead of inventing a specific identity or exact event facts. Treat source-aware or source-backed as a request for careful, credible subject matter, not prose about sources or research work.

Partition the outcome into distinct useful image-plus-content units. For each unit separate:
\begin{itemize}[leftmargin=1.4em, itemsep=0.15em, topsep=0.3em]
  \item \texttt{image\_subject}: one concrete physical subject or state that a single independent
  image can directly depict in one coherent view;
  \item \texttt{content\_goal}: one short reader-facing question asking what useful explanation,
  interpretation, comparison, instruction, chronology, or decision the adjacent prose must answer.
  Ask about the subject itself.
\end{itemize}

Normally create one unit for each distinct requested physical subject, showing a view that best supports all of its requested properties; put the properties and comparison criteria together in that unit's \texttt{content\_goal}. Create another unit for the same subject only when the Task explicitly requires a real change of state, time, or location. A different evaluation criterion, viewpoint, trade-off, comparison, recommendation, or prose section never creates a unit. Never create a summary, diagram, plan, split view, collage, or information-display image. Audience, later use, and presentation format do not add subjects.

Interpret the requested action literally. The endpoint is the earliest completed requested outcome or the last event explicitly stated in a narrative. For set up, arrange, or organize, operating or demonstrating the ready items is a later task. In a narrative, put the immediate reaction to the last event in that event's content rather than adding an epilogue unit.

When the Task's main operation is comparing named alternatives, create exactly one unit for each alternative unless the Task explicitly requests a real before/after or other state change within an alternative. Each unit's one view must show that alternative in the use context that exposes the most requested properties; its \texttt{content\_goal} carries all criteria, trade-offs, and the decision. Before returning, verify that unit count equals alternative count unless that exception applies. For comparison tasks, an \texttt{image\_subject} containing two or more named alternatives is invalid, and a step whose only purpose is to synthesize or conclude the comparison is invalid.
Put synthesis and the decision in the \texttt{content\_goal} of the last alternative's existing unit.

An interleaved report requires at least two distinct, useful image-plus-content units, so output at least two semantic steps. Every step must make a different task-relevant contribution. Do not add steps merely to satisfy the minimum.

Output exactly:

\begingroup
\ttfamily\raggedright
\setlength{\parindent}{0pt}

\noindent\{"title":"...",\\
"endpoint":"earliest completed outcome or last explicit event",\\
"steps":[\{"step\_index":1,\\
"image\_subject":"one concrete visible subject or state",\\
"content\_goal":"short reader-facing question"\}]\}

\par
\endgroup

The top-level object has exactly these three keys. Every object inside steps has exactly three keys:
\texttt{step\_index}, \texttt{image\_subject}, and \texttt{content\_goal}.
}{prompt:plan1}

\myparagraph{Plan \#2: Text and Image Planning.}
Generates the reader-facing prose and image-generation instruction separately for each unit, together with a shared visual style across the output.

\prompt{Text and Image Planning}{%

Task request: \texttt{\{task\}}

Write concise finished prose for the reader. Directly answer each \texttt{content\_goal} using task details, ordinary knowledge, and plausible scene detail. Keep an unspecified subject at the category level instead of inventing a specific identity or exact event facts. Discuss the subject, not the writing process. Give useful information directly; do not assign research to the reader.

\noindent Frozen scope and sequence: \texttt{\{steps\_summary\}}\\
Frozen endpoint: \texttt{\{endpoint\}}\\
Frozen step indexes: \texttt{\{step\_indexes\}}

Choose one shared \texttt{visual\_style} for rendering treatment only: medium, color treatment, lighting character, and rendering finish. Keep subjects, actions, locations, viewpoint, composition, and prose voice in the individual steps. Follow an explicitly requested style; otherwise use a restrained treatment appropriate to the subject.

Return one object for each frozen step index, in order. The complete task defines global scope and each frozen unit separates its visible subject from its prose contribution. Apply every relevant task constraint without changing the frozen sequence.

For each step output:
\begin{itemize}[leftmargin=1.4em, itemsep=0.15em, topsep=0.3em]
  \item the same \texttt{step\_index}; do not repeat Plan1 fields in the output;
  \item \texttt{image\_prompt}: one complete, self-contained rendering instruction for the
  observable subject, state, relations, and context required by this contribution;
  \item \texttt{content}: finished report prose carrying useful explanation that need not be
  visible in pixels.
\end{itemize}

Name required visible elements concretely rather than replacing them with generic collective terms.
Include only entities whose visible presence follows from the requested contribution. Keep one coherent physical view per image. Depict the frozen \texttt{image\_subject} rather than a container for presenting information about it, unless that container is itself the requested subject.

Do not turn a comparison or synthesis into a split composition, collage, multi-panel sequence, overlay, diagram, plan view, map, report page, label panel, or information display. The separate images provide the concrete subjects; content carries relations across them. Across the content fields, cover the complete task.

Output schema:

\begingroup
\ttfamily\raggedright
\setlength{\parindent}{0pt}

\noindent\{"visual\_style":\{"medium":"...",\\
"color\_treatment":"...",\\
"lighting\_character":"...",\\
"rendering\_finish":"..."\},\\
"steps":\{step\_semantic\_entries\}\}

\par
\endgroup
}{prompt:plan2}

\myparagraph{Plan \#3: Selective Visual Conditioning.}
Constructs an optional bank of up to two reusable visual references based on the planned image prompts and assigns relevant references to individual units.

\prompt{Selective Visual Conditioning}{%

\noindent Task: \texttt{\{task\}}\\
Frozen semantic subjects: \texttt{\{semantic\_subjects\}}\\
Rendered Steps: \texttt{\{steps\_summary\}}

Select zero to two isolated identity references. Before writing JSON, silently enumerate every concrete person, object, vessel, tool, carrier, and support explicitly visible in each rendered Step, then group only occurrences established as the same physical individual. Do not group repeated members of a category.

Treat identity as established when an ordered construction or setup keeps an explicitly visible stable core in later Steps, even if later wording omits the phrase ``the same'': a completed artifact retains its visibly described support frame, and a cumulative arrangement retains a named tool that is still explicitly shown. Do not infer identity from category repetition across independent scenes.

For each candidate, silently perform this counterfactual test: imagine supplying one isolated reference image unchanged to every proposed Step. Reject the candidate if any assigned Step needs the candidate itself to have a different outline, topology, fold or crease pattern, construction, surface state, or identity. Changes only to contents, attachments, pose, action, lighting, camera, or surroundings do not alter identity when the unchanged core remains visible. A container can therefore persist while its contents change, and a support frame can persist while panels are attached; a sheet being cut or folded cannot persist as one unchanged visual reference.

A stable identity shown in only one Step may optionally receive a reference when its specific identity matters. Every candidate shown in two or more Steps that passes the counterfactual test must enter the bank and must be routed to every Step where it is explicitly visible. If more than two pass, keep the two with greatest Step coverage, then greatest importance to the main artifact.
Return an empty bank only when no candidate passes. Do not create references for generic category members, incidental props, scenes, backgrounds, landforms, layouts, spatial relations, routes, or information carriers. Output only the final schema; do not expose the test or reasoning. A reference prompt depicts only the invariant target in isolation, without a changing state, action, scene, or background.

Each item contains only fields that the image runtime consumes:
\begin{itemize}[leftmargin=1.4em, itemsep=0.15em, topsep=0.3em]
  \item \texttt{key}: stable identifier;
  \item \texttt{type}: reference;
  \item \texttt{image\_prompt}: a self-contained image of only the stable reusable target.
\end{itemize}

Every selected key must be referenced by at least one \texttt{step\_conditions} entry. Include every \texttt{step\_N} and use \texttt{[]} when no reference applies. Each \texttt{step\_conditions} array contains only key strings from \texttt{condition\_bank.items}, never image prompts or descriptions.
When \texttt{items} is empty, every \texttt{step\_conditions} array must be \texttt{[]}.

Default output when there is no qualifying cross-step identity reference:

\begingroup
\ttfamily\raggedright
\setlength{\parindent}{0pt}

\noindent\{"condition\_bank":\{"items":[]\},\\
"step\_conditions":\\
\{step\_condition\_entries\}\}

\par
\endgroup

Only when a qualifying reference exists, an item has exactly this shape:

\begingroup
\ttfamily\raggedright
\setlength{\parindent}{0pt}

\noindent\{"key":"stable\_identifier",\\
"type":"reference",\\
"image\_prompt":"self-contained image of only the stable reusable target"\}

\par
\endgroup
}{prompt:plan3}

\myparagraph{Plan \#4: Visual Verification Checklist.}
Derives one or two directly observable visual checks per unit. These checks concern visible objects, actions, and states in the corresponding image.

\prompt{Visual Verification Checklist}{%

Checks must rely on directly visible non-text evidence.
Do not require reading or recognizing text, symbols,
labels, numbers, screens, or inscriptions.

Steps:\\
\texttt{\{steps\_summary\}}

Give each step 1--2 checks that a reviewer can verify from that image.
Check the assigned observable evidence and any exact visible count required
by the step. Output only \texttt{step\_N} checks.

\begingroup
\ttfamily\raggedright
\setlength{\parindent}{0pt}

\noindent\{"reviewer\_checklist":\\
\hspace*{2em}[\{step\_checklist\_entries\}]\}

\par
\endgroup
}{prompt:plan4}

\myparagraph{Execution and Refinement.}
Reference images are generated before the unit images.
Each unit is rendered from its image prompt and the shared visual style, together with any assigned reference images that the backbone accepts. The reviewer evaluates the generated image against its checklist and returns a \texttt{PASS} or \texttt{FAIL} result for each criterion. Failed criteria guide regeneration of the corresponding image, while successful units remain unchanged. The final output pairs the selected images with the prose generated in Plan 2.

\subsection{Stage Schemas and Runtime Settings}

Table~\ref{tab:apx-schemas} summarizes the inputs and outputs of each stage. Earlier planning outputs remain fixed, allowing each stage to be validated independently.

\begin{table}[h]
\centering
\small
\caption{
\textbf{UniWeaver organizes multimodal generation through four structured planning stages and visual verification.}
The table summarizes the inputs and outputs of each stage and the image reviewer.
}
\label{tab:apx-schemas}
\begin{tabularx}{\linewidth}{@{}lXX@{}}
\toprule
Stage & Input & Output \\
\midrule
$P_1$
& User request
& Title, task endpoint, and unit-level visual subjects
and content goals \\

$P_2$
& Request and $P_1$
& Shared visual style, reader-facing prose,
and image prompts \\

$P_3$
& Request and planned image prompts
& Reference bank (0--2 images) and
unit-specific reference assignments \\

$P_4$
& Planned image prompts
& Observable visual checks for each unit \\

Review
& Generated image and checklist
& Per-criterion \texttt{PASS}/\texttt{FAIL}
with feedback \\
\bottomrule
\end{tabularx}
\end{table}

Each report uses at most two reference images.
For backbones supporting only one conditioning image, only the first routed reference is supplied.
All planning stages use greedy decoding at temperature $0$. The maximum output lengths for P1--P4 are $2{,}048$, $2{,}560$, $768$, and $1{,}024$ tokens, respectively. The reviewer returns one judgment per checklist item.
Training trajectories must satisfy the prescribed schemas and task-consistency checks.

\section{Benchmark Construction}
\label{sec:apx-benchmark}

\subsection{Design and Composition}

UniWeaver-Bench contains 120 goal-only tasks spanning 20 task families across four categories, with six held-out tasks per family and a balanced difficulty distribution of 40 easy, 40 medium, and 40 hard tasks.
The task families follow the taxonomy of UniWeaver-10K, as illustrated in Figure~\ref{fig:dataset_statistics}(a), while benchmark tasks are held out from training. The four categories group the 20 families as follows:
\begin{itemize}[leftmargin=1.4em, itemsep=0.1em, topsep=0.2em]
  \item \textit{Procedures \& Instructions}: Recipe \& Food
  Preparation, Craft \& DIY, Assembly \& Physical Setup,
  Care \& Maintenance, and Before-and-After Transformation.
  \item \textit{Comparison \& Recommendation}: Product / Object
  Comparison, Place / Experience Comparison, Product Use \&
  Lifestyle Guide, Fashion \& Styling, and Interior \&
  Lifestyle Inspiration.
  \item \textit{Observation \& Description}: Photography / Visual
  Showcase, Nature / Species / Environment, Architecture \& Design
  Style, Seasonal / Atmosphere / Environmental Change, and Travel
  \& Place Guide.
  \item \textit{Explanation \& Storytelling}: Collection / Visual
  Catalog Report, Object History \& Evolution, Culture \&
  Tradition, Event / Experience Story, and Narrative / Slice of Life.
\end{itemize}

The benchmark therefore evaluates generalization to unseen requests within familiar task families.
Final scores are aggregated with equal weight across the 20 families, following Section~\ref{sec:3.3}.

\subsection{Evaluator-Only Fields and Separation}

Each benchmark task includes three evaluator-only fields: task family, difficulty and hidden evaluation reference.
The evaluated method receives only the task identifier and user request, while evaluation metadata and hidden references remain inaccessible during generation.
The hidden references specify task-specific content requirements and observable visual evidence.

To prevent training--evaluation overlap, we compare UniWeaver-10K against the frozen 120-task benchmark using task identity and normalized prompt hashes.
No exact matches were detected among the 10,000 retained training tasks. This filtering prevents identified task and prompt duplication.

We conducted a content-screening pass over all 120 benchmark tasks, including both the user-facing prompts and their hidden evaluation-reference fields.
The procedure applied deterministic keyword and pattern filters to identify objectionable content. Neither offensive content nor personally identifying information was detected.

\section{Evaluation Protocol}
\label{sec:apx-eval}

UniWeaver-Bench evaluates the final reader-facing text and generated images, rather than internal planning trajectories. A deterministic validity check requires a nonempty sequence of ordered text-image units, each containing reader-facing content paired with an existing, non-reused image. The check operates only on the final output. Invalid reports do not enter $\bar{Q}_f$ and are penalized through $\sqrt{r_f}$.

Valid reports are evaluated using a task-specific rubric across four dimensions: four to five for text coverage (TC), two for writing quality (WQ), two to three for visual coverage (VC), and one per text-image unit for image-text alignment (ITA).
Each criterion is rated as \emph{not met}, \emph{partially met}, or \emph{met}, corresponding to $0$, $0.5$, and $1$, respectively.
Table~\ref{tab:apx-report} summarizes the evidence used for each dimension.

Each method is run once on all 120 tasks using a fixed base generation seed of 42. The two repeated judge calls per criterion are used only for within-judge aggregation.

\begin{table}[h]
\centering
\small
\caption{
\textbf{UniWeaver-Bench evaluates final outputs using a task-specific rubric across four dimensions.}
The table specifies the number of criteria and evaluation evidence for each dimension.
}
\label{tab:apx-report}
\begin{tabularx}{\linewidth}{@{}lcX@{}}
\toprule
Dimension & \# Criteria & Evaluation Evidence \\
\midrule
TC & 4--5 & User request, hidden coverage requirements,
and final text \\
WQ & 2 & User request and final text \\
VC & 2--3 & User request, hidden visual requirements,
and generated images \\
ITA & one per unit & User request, final text, and
corresponding images \\
\bottomrule
\end{tabularx}
\end{table}

\begin{figure*}[t]
 \centering
 \includegraphics[width=\textwidth]{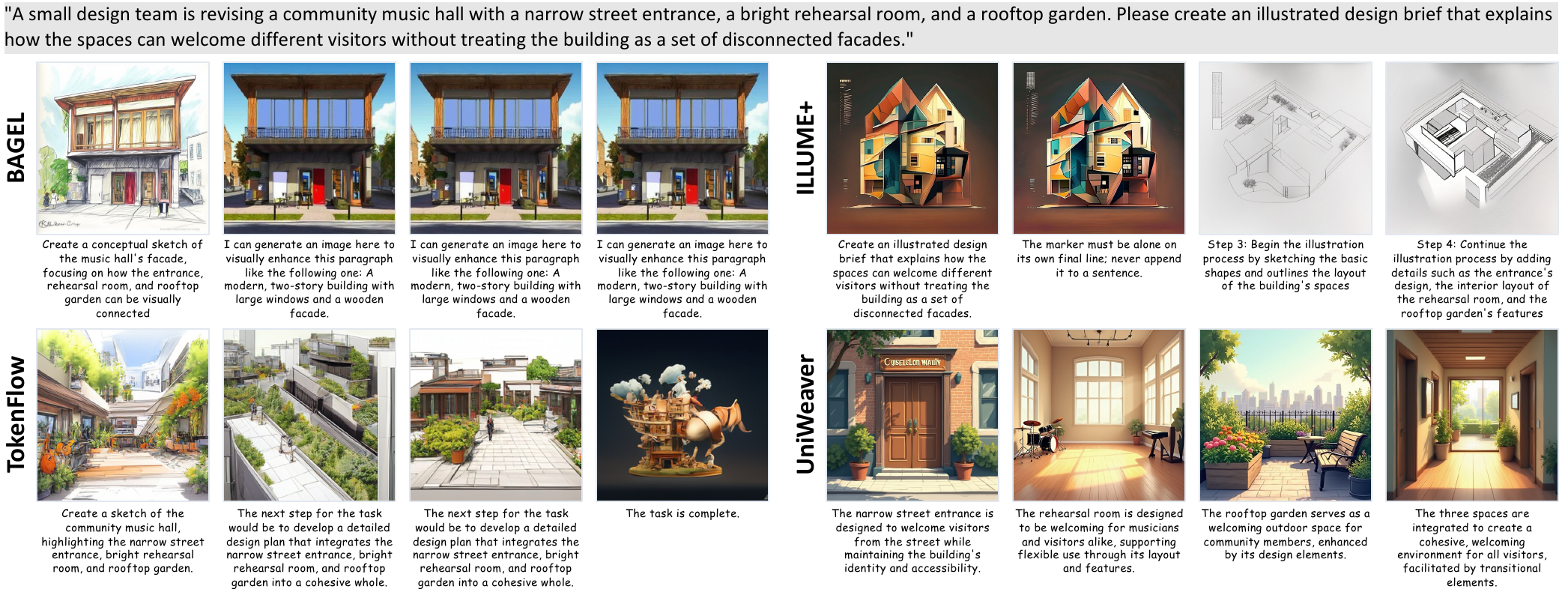}
    \caption{
\textbf{Additional qualitative comparison on the architectural design task.}
UniWeaver produces coherent, task-specific descriptions and corresponding visuals covering the entrance, rehearsal room, rooftop garden, and their integration, while baseline UMMs exhibit repetitive content or incomplete task coverage.
}
\vspace{-3mm}
    \label{fig:appendix_fig}
\end{figure*}

\subsection{Judging and Aggregation}

We independently evaluate valid reports using three judges: Kimi K3 \citep{DBLP:journals/corr/abs-2607-24653}, DeepSeek-V4.1-Flash \citep{DBLP:journals/corr/abs-2609-19969}, and GLM-5.3-Flash \citep{DBLP:journals/corr/abs-2602-15763}. Each judge evaluates one task at a time at temperature $0$, with two repeated calls per criterion. The median rating is used as the within-judge result.

Cross-judge aggregation is conservative. When the three medians agree, that rating is used. A one-tier disagreement is resolved by taking the lower rating, and a disagreement spanning more than one tier is scored as $0$, the lowest tier.

Text and visual evaluations use separate fixed system instructions. Judges receive only the task-specific materials and evaluation references. Each judgment returns structured JSON containing criterion-level ratings and brief supporting observations. Visual evaluations rely on directly observable image evidence, and rendered text is not treated as reliable visual evidence unless it is itself the evaluation target.

Panel ratings are normalized to quarter steps between $0$ and $1$, where a criterion whose three medians span more than one tier scores $0$. Task quality is calculated using the weighted criteria defined in Section~\ref{sec:3.3}. Within each task family, $Q_t$ is averaged over gate-passing tasks and multiplied by the square root of the gate pass rate. Final scores are obtained by equally averaging the 20 family scores and are reported on a $0$--$100$ scale.

\section{Training and Model Integration}
\label{sec:apx-training}

We apply LoRA to BAGEL's understanding decoder (14.22B total parameters and 0.08B trainable parameters) while keeping the image-generation experts and vision encoder frozen. UniWeaver-10K contains 10,000 trajectories, each expanded into four planning-stage training examples. With two examples packed per optimization step, one epoch uses 20,000 steps; we train for two epochs, for 40,000 steps in total. The run uses one NVIDIA GB10 GPU, processes 101.3M tokens in about 35 GPU-hours, and peaks at 79.5 GiB of GPU memory. These timings exclude judge charges. Table~\ref{tab:apx-training} summarizes the optimization settings. All reported UniWeaver-SFT results use the final checkpoint.

\begin{table}[h]
\centering
\small
\caption{
\textbf{UniWeaver-SFT adapts BAGEL's understanding decoder using LoRA.}
The table summarizes the key training and optimization settings.
}
\label{tab:apx-training}
\begin{tabularx}{\linewidth}{@{}lX@{}}
\toprule
Setting & Value \\
\midrule
LoRA (rank / alpha / dropout) & $32$ / $64$ / $0.05$ \\
Optimizer & AdamW, $\beta=(0.9,0.95)$,
$\epsilon=10^{-8}$, weight decay $0.01$ \\
Learning rate & $5\times10^{-5}$ \\
Schedule & $200$-step warm-up, cosine decay \\
Training steps & $40{,}000$ \\
Gradient clipping / seed & $1.0$ / $42$ \\
Max. sequence length & $4{,}096$ tokens \\
\bottomrule
\end{tabularx}
\end{table}

\section{Details of Ablation Studies}
\label{sec:apx-ablation}

Table~\ref{tab:apx-ablation} specifies the implementation of the seven variants
in Table~\ref{tab:ablation}. Each variant modifies only the indicated component,
preserves the original planning-stage order, and keeps every remaining stage fixed. In particular, ``References Fixed before the Text'' changes only which upstream fields are visible when $P_3$ selects references; it retains the stage's timing and output interface. This isolates the effect of information order from changes in reference availability or per-step routing.

All seven variants start from the same base BAGEL model and use identical
evaluation settings; none of them loads the SFT checkpoint. They also share the
same benchmark tasks, non-ablated prompts, decoding configuration,
image-generation settings, and judge protocol.

\begin{table}[t]
\centering
\fontsize{9pt}{10.5pt}\selectfont
\setlength{\tabcolsep}{2pt}
\renewcommand{\arraystretch}{1.08}
\caption{
\textbf{Seven ablation variants isolate individual components of UniWeaver's planning workflow.}
The table specifies the modifications to $P_1$, $P_2$, and $P_3$.
}
\vspace{-2mm}
\label{tab:apx-ablation}
\begin{tabularx}{\columnwidth}{@{}>{\raggedright\arraybackslash}p{0.46\columnwidth}>{\raggedright\arraybackslash}X@{}}
\toprule
Ablation & Modification \\
\midrule
\multicolumn{2}{@{}l@{}}{%
    \begingroup
    \setlength{\fboxsep}{1pt}%
    \colorbox{gray!20}{%
        \makebox[\dimexpr\columnwidth-2\fboxsep\relax][l]{%
            \textbf{$\boldsymbol{P_1}$: Task Scope}%
        }%
    }%
    \endgroup
} \\
\hspace{0.6em}w/o Task Goal
& Removes the \texttt{endpoint} field from $P_1$ \\
\hspace{0.6em}w/o Per-step Content Goal
& Removes unit-level content goals while retaining visual subjects \\
\multicolumn{2}{@{}l@{}}{%
    \begingroup
    \setlength{\fboxsep}{1pt}%
    \colorbox{gray!20}{%
        \makebox[\dimexpr\columnwidth-2\fboxsep\relax][l]{%
            \textbf{$\boldsymbol{P_2}$: Text and Image Planning}%
        }%
    }%
    \endgroup
} \\
\hspace{0.6em}w/o Shared Visual Style
& Removes the shared visual style from $P_2$ \\
\hspace{0.6em}w/ Merged Image Prompt and Content
& Uses one field for both reader-facing prose and image-generation instructions \\
\multicolumn{2}{@{}l@{}}{%
    \begingroup
    \setlength{\fboxsep}{1pt}%
    \colorbox{gray!20}{%
        \makebox[\dimexpr\columnwidth-2\fboxsep\relax][l]{%
            \textbf{$\boldsymbol{P_3}$: Visual References}%
        }%
    }%
    \endgroup
} \\
\hspace{0.6em}References Fixed before the Text
& Selects references using only $P_1$, without access to~$P_2$~outputs \\
\hspace{0.6em}w/o Per-step Reference Routing
& Retains the reference bank but removes unit-specific routing \\
\hspace{0.6em}w/o All Reference Images
& Disables reference image generation and conditioning \\
\bottomrule
\end{tabularx}
\vspace{-2mm}
\end{table}

\section{Additional Qualitative Results}
\label{sec:apx-additional-qualitative}

Figure~\ref{fig:appendix_fig} presents an architectural design task that requires integrating a street entrance, rehearsal room, and rooftop garden into a coherent illustrated brief. Unlike the instructional tasks in the main text, this case examines how models organize multiple functional spaces and explain their relationships.

BAGEL repeatedly generates similar exterior views,
TokenFlow produces repetitive planning-oriented content, and ILLUME+ exposes intermediate generation instructions instead of complete reader-facing descriptions.
In contrast, UniWeaver provides distinct, task-relevant text-image units for each space and concludes by explaining their integration into a unified design.

The architectural brief is a useful stress test because each functional space must remain individually identifiable while also contributing to one coherent design. Methods that generate units independently can produce plausible local views without preserving their cross-space relationships, whereas methods with stronger sequential dependence can collapse multiple spaces into near-duplicate views. The example therefore exposes the balance UniWeaver targets: sufficient independence to cover distinct content units, together with enough shared conditioning to preserve a coherent overall result.

\end{document}